\documentclass{article} 
\usepackage{iclr2025_conference,times}

\usepackage{amsmath,amsfonts,bm}

\def\eqref#1{equation~\ref{#1}}

\def\1{\bm{1}}

\DeclareMathAlphabet{\mathsfit}{\encodingdefault}{\sfdefault}{m}{sl}
\SetMathAlphabet{\mathsfit}{bold}{\encodingdefault}{\sfdefault}{bx}{n}

\usepackage{hyperref}
\usepackage{url}
\usepackage{graphicx}
\usepackage{booktabs}

\title{Beyond Glucose-Only Assessment: Advancing Nocturnal Hypoglycemia Prediction in Children with Type 1 Diabetes}

\author{Marco Voegeli\thanks{Equal contribution. Correspondence to \texttt{slaguna@inf.ethz.ch}} \\
ETH Zurich\\
\And
Sonia Laguna\footnotemark[1] \\
ETH Zurich\\
\And
Heike Leutheuser \\
University of Bayreuth \\
\And
Marc Pfister \\
University Children’s Hospital Basel \\
\And
Marie-Anne Burckhardt \\
University Children’s Hospital Basel \\
\And
Sara Bachmann \\
University Children’s Hospital Basel \\
\And
Julia E. Vogt \\
ETH Zurich \\
}

\iclrfinalcopy 
\begin{document}

\maketitle

\begin{abstract}

The \textit{dead-in-bed} syndrome describes the sudden and unexplained death of young individuals with Type 1 Diabetes (T1D) without prior long-term complications. One leading hypothesis attributes this phenomenon to nocturnal hypoglycemia (NH), a dangerous drop in blood glucose during sleep. This study aims to improve NH prediction in children with T1D by leveraging physiological data and machine learning (ML) techniques. We analyze an in-house dataset collected from 16 children with T1D, integrating physiological metrics from wearable sensors. We explore predictive performance through feature engineering, model selection, architectures, and oversampling. To address data limitations, we apply transfer learning from a publicly available adult dataset. Our results achieve an AUROC of 0.75 $\pm$ 0.21 on the in-house dataset, further improving to 0.78 $\pm$ 0.05 with transfer learning. This research moves beyond glucose-only predictions by incorporating physiological parameters, showcasing the potential of ML to enhance NH detection and improve clinical decision-making for pediatric diabetes management.

\end{abstract}

\section{Introduction}
\label{sec:introduction}

Type 1 Diabetes (T1D) is a chronic autoimmune disease characterized by
the destruction of pancreatic beta cells ~\citep{t1d_paper}. In 2021, approximately 8.4 million individuals worldwide had T1D, 18\% being under 20 years. By 2040, the number of cases is projected to increase by 60-107\%, particularly in low-income and lower-middle-income countries~\citep{gregory2022global}. 
A major complication of insulin-treated diabetes is hypoglycemia, which occurs when blood glucose levels drop below 70 mg/dL (3.9 mmol/L)~\citep{hypoglycemia}. Hypoglycemia occurring at night, mainly during sleep, is known as nocturnal hypoglycemia (NH) and poses a severe threat
to individuals with T1D~\citep{danger_hyp, danger_nocthyp}. Because individuals are often unaware of hypoglycemic events while asleep, they are unable to take corrective actions in real time. Beyond the immediate physiological risks, NH can impair sleep quality, reduce daytime cognitive function, and increase the likelihood of cardiovascular complications, psychological distress, and fear of future episodes~\citep{allen2003nocturnal, Brod2012}. An early-warning system for NH could help mitigate these risks by allowing individuals to take necessary precautions in advance.

Glycemic variability, a key factor influencing NH, is affected by several daily lifestyle habits, including physical activity~\citep{EffectofPhysicalActivityonGV}. Managing glycemic variability in children is particularly challenging due to the physiological changes during puberty and their limited understanding of the topic \citep{Nadella2017, Franzese2004}. 
Wearable health monitoring devices have gained widespread adoption~\citep{piwek2016rise, lu2020wearable}, presenting an opportunity to integrate physiological data into NH prediction models. By leveraging these devices in combination with machine learning (ML) models, the prediction and prevention of NH could be significantly improved, reducing life-threatening complications.

ML has shown promise in the medical domain for predictive modelling, diagnostics, and automation of complex decision-making tasks~\citep{Krizhevsky2012ImageNetCW, he2015deep}. However, challenges such as data sparsity, limited patient-specific training data, and out-of-domain distributions complicate the development of robust predictive models~\citep{javaid2021significance, araujo2016using}. To address these challenges, this study employs tailored ML techniques to predict NH in children with T1D.

Overall, our main contributions are: (i) We introduce a novel NH prediction approach in children, going beyond traditional glucose-only methods by integrating physiological signals from wearable sensors and focusing on a challenging long prediction horizon, not generally tackled in the literature. (ii) We explore advanced feature selection, tailored preprocessing, and optimized model architectures to tackle the challenges of high-feature, low-cardinality data. (iii) We demonstrate the power of transfer learning by effectively leveraging adult data to enhance predictive performance for pediatric diabetes management.

\section{Related Work}
\label{sec:rel_work}
Several studies have explored ML approaches for predicting NH, leveraging different datasets, prediction horizons, and modelling techniques. \citet{vuExtendedPred2020} investigated NH classification for two prediction windows: 0:00-3:00 am and 3:00-6:00 am, achieving AUROC scores of 0.90 and 0.75, respectively. Their model, a Random Forest Classifier (RFC), was trained on a large dataset comprising 1 million continuous glucose monitoring (CGM) data points from adults aged 45.34$\pm$16.38 years. \citet{lopezCBigDNH} focused on predicting minimum nocturnal glucose concentration across an entire night using a dataset of 22,804 nights from donors aged 31$\pm$19 years. The authors extracted features from CGM, insulin intake, and meal data and employed a Support Vector Regressor (SVR) to estimate glucose concentration. The model then predicted NH events, achieving 94.1\% accuracy in correctly identifying NH nights and an AUROC score of 0.86 (95\% CI, 0.75–0.98). \citet{berikov} took a different approach, using a dataset of 36,900 CGM data points from adults aged 18–70 years, along with 23 clinical and laboratory parameters. They employed an RFC model to predict NH at shorter prediction horizons of 15 and 30 minutes, achieving AUROC scores of 0.97 and 0.942, respectively. This study incorporated glucose metric extraction and additional physical parameters influencing NH risk. Lastly, \citet{Bertachi2020} examined NH prediction using physiological metrics from wearable sensors in addition to CGM data. The study collected data over 12 weeks from 10 adult participants, yielding approximately 840 nights of data. The authors trained an SVM model to predict NH with a 6-hour prediction horizon, beginning at sleep onset, achieving sensitivity and specificity scores of 78.75\% and 82.15\%, respectively. Despite the smaller dataset, our work stands out by integrating physiological features from wearable devices and exploring transfer learning across pediatric and adult datasets, providing a more robust evaluation framework and addressing key gaps in NH prediction research.


\section{Study Framework: Datasets for NH} %

\subsection{In-house Dataset}
\label{sec:diacamp}
\paragraph{Study Setup}

The in-house dataset originates from a one-week sports day camp for children with T1D, approved by the local ethics committee. The study ran from 7:00 am on day 1 to 10:00 am on day 7, with pediatric endocrinologists supervising from 9:00 am to 5:00 pm. Activities, insulin treatment, and nutrition throughout the study were standardized. The first day included climbing, while days 2–6 featured structured sports. The final day, mainly concluding the study, was excluded from the analysis due to missing overnight data.

\paragraph{Participants}
16 children aged 7–16 years, diagnosed with T1D for at least six months, participated. They used either multiple daily injections (MDI) or continuous subcutaneous insulin infusion (CSII) for insulin therapy. Written informed consent was obtained from children and/or caregivers before the study. Finally, data from 11 children were used; the remaining 5 were excluded due to recording errors.

\paragraph{Hardware}
Devices used for this study consisted of a physiological wearable sensor Everion (Biofourmis, Boston,
US) and a continuous subcutaneous glucose sensor. The Everion devices continuously recorded vital signs throughout the study (described here: \ref{paragraph:feature_A}) and were typically charged or replaced each morning when the children arrived at the camp. Glucose was measured continuously by a continuous subcutaneous glucose sensor; low glucose values were confirmed by a fingerprick (self-monitoring blood glucose SMBG) measurement. These recordings throughout the study were stored as distinct databases: an Everion database, consisting of the vital signs recorded by the Everion sensor; and a glucose database, storing the glucose readings, CGM, and SMBG. The Everion sensor is a CE-certified research device with a sampling rate of 1 Hz that captures 22 vital signs in real time, with corresponding quality measures. The Everion sensor was fitted on the upper part of the participant's arm (right or
left). The glucose sensors for this study are intermittently scanned continuous glucose monitoring (isCGM), Freestyle libre 2 (Abbott Diabetes Care Inc., Alameda, US),
CGM, Dexcom (Dexcom, San Diego, US) or Guardian 3 (Minimed Medtronic,
Northridge, US) with a sampling rate of 5 minutes for the CGM devices and 15 minutes for the isCGM system.

\paragraph{Manual records}
The glucose dataset was completed with the SMBG records. The SMBG measurements were taken each time hyperglycemic or hypoglycemic symptoms were observed, i.e. sensor measurements were below 3.9 mmol/l or above 15 mmol/l, before and after physical activity and hourly during physical activity. Insulin doses in type, time, and units, carbohydrate intake, type and duration of physical activity, symptoms of
hypoglycemia, and SMBG were noted in a
logbook by the study team. The children continued the measurements and logbook entries at home in the evenings, nights, and mornings. Children's metadata, including morphological information such as age, weight, height, and body mass index (BMI) for each participant, were also recorded. 

 \paragraph{Features}
 \label{paragraph:feature_A}
The dataset encompasses three distinct categories of features: time-dependent features detailed in Section \ref{apd:feat_set}, logbook-recorded features presented in Table~\ref{tab:logbook}, and patient-specific metadata attributes presented in Table ~\ref{tab:meta_data}.
\vspace*{-0.5cm}
\begin{table}[h!]
\caption{List and description of features in the logbook dataset.}
\begin{center}
\begin{tabular}{p{4.65cm}p{5.4cm}}
\toprule
\textbf{Logbook Feature} & \textbf{Description} 
\\ \midrule
Date & Date of the data entry \\ 
Time & Time of the data entry \\ 
Blood Sugar & Glucose level in the blood \\ 
Sensor Glucose & Sensor measured glucose level\\
SGL Trend & Time trend of sensor glucose level \\ 
Basal Insulin & Baseline insulin dose \\ 
Rapid-Acting Insulin Meals & Insulin dose for meal times \\ 
Rapid-Acting Insulin Correction & Dose to correct high blood sugar \\ 
Carbohydrates (g) Mixed & Mixed carbohydrates intake in grams\\
Carbohydrates (g) Fast & Fast-absorbing carbohydrates intake\\
Carbohydrates (g) Slow & Slow-absorbing carbohydrates intake \\
Hypo Correction (yes/no) & Whether a hypo correction was made \\
Type of Carbohydrates & The type of carbohydrates consumed \\
Duration of Exercise & Duration of physical activity \\
Duration of Exercise (estimate) & Estimated duration of physical activity \\
Type of Sport/Activity & Type of physical activity performed \\
Hypo Symptoms & Presence of hypoglycemia symptoms \\ 
Remarks & Additional notes \\ 
\bottomrule
\end{tabular}
\end{center}
\label{tab:logbook}
\vspace*{-0.8cm}
\end{table}

\subsection{OhioT1DM Dataset}
\label{ohio}
The OhioT1DM dataset by \citet{marling2020ohiot1dm} is a comprehensive and well-curated dataset specifically designed for research in T1D management. It includes detailed physiological and behavioural data collected from adults with T1D, providing valuable insights for developing and testing predictive models and treatment strategies.

\paragraph{Study Setup}
The dataset comprised two 8-week studies, one released in 2018 and one in 2020. The studies used different sensor bands with varying sampling rates and sensors.

\paragraph{Participants}
The dataset released in 2018 involved six participants, two males and four females, aged 20 to 40, and thus in 2020, six participants, five males and one female, aged 20 to 80.

\paragraph{Hardware}
The participants wore Medtronic 530G or 630G insulin pumps and Medtronic Enlite CGM sensors throughout the 8-week data collection. They reported life-event data via a custom smartphone app and physiological data from a fitness band. For the 2018 dataset, Basis Peak fitness bands were used. The six individuals used from the 2020 set wore the Empatica Embrace device. 

 \paragraph{Features}

The OhioT1DM dataset's full list of features is listed in \citet{marling2020ohiot1dm}. This study worked on the following features: Insulin type, glucose level (CGM data), finger stick (blood glucose values obtained through self-monitoring by the patient), hypo event (time of self-reported hypoglycemic episode), basis heart rate (heart rate, aggregated every 5 minutes), basis GSR (galvanic skin response, aggregated every 5 minutes), basis steps (step count, aggregated every 5 minutes), basis sleep (times when the sensor band reported that the subject was asleep), and acceleration (magnitude of acceleration, aggregated every 1 minute). A 5-minute aggregation of heart rate data is only available when participants wore the Basis Peak band (2018 cohort).

\subsection{Labels}
Labels for both datasets were calculated using overnight CGM and SMBG recordings (10 pm - 7~am). A night was classified as hypoglycemic if it met either of the following criteria: (1) CGM readings fell below the hypoglycemic threshold (3.9 mmol/L) for at least 15 consecutive minutes, or (2) any SMBG measurement was below 3.9 mmol/L. In the OhioT1DM dataset, CGM represents glucose levels, while SMBG refers to the fingerstick measurements.


\vspace*{-0.05cm}
\section{Methods} %
\subsection{Data Preprocessing} 

Prior to feeding the data to the predictive models, we preprocess it for homogeneity. The in-house dataset and OhioT1DM contained 60 and 308 labels, respectively.
The sensors recorded data 24 hours a day, however for the in-house dataset the Everion devices were often swapped out in the mornings due to battery levels. This meant that a substantial amount of the recordings between 7am and 10am were missing. To reduce bias and minimize missing data, we restricted our analysis to the period between 10am and 10pm. This approach excludes the 7am to 10am window, which contained minimal useful information, ensuring that data closer to the prediction horizon—critical for accurate predictions—are prioritized.
To further remove missing values from the sensor recordings we applied thresholding for each signal, the thresholds were taken from the technical ranges specified in Everion manual. The remaining missing values were substituted with zeros. 
Imputation methods such as polynomial and linear interpolation, excluding days with excessive missing values, and forward fill were considered. However zero imputation yielded the best results for our model.

When working on medical datasets, class imbalance is a common problem~\citep{suresh2023imbalanced}. Especially for our study, the severe health risks posed by false negatives made us explore the distribution of our labels. The in-house dataset has a label imbalance ratio of 1:5.5 for hypoglycemic nights to normal nights, while the OhioT1DM dataset has a ratio of 1:2.1 respectively. Hence, we had to apply modifications to avoid overfitting.

\begin{figure*}[h!]
    \centering
    \includegraphics[width=0.6\textwidth]{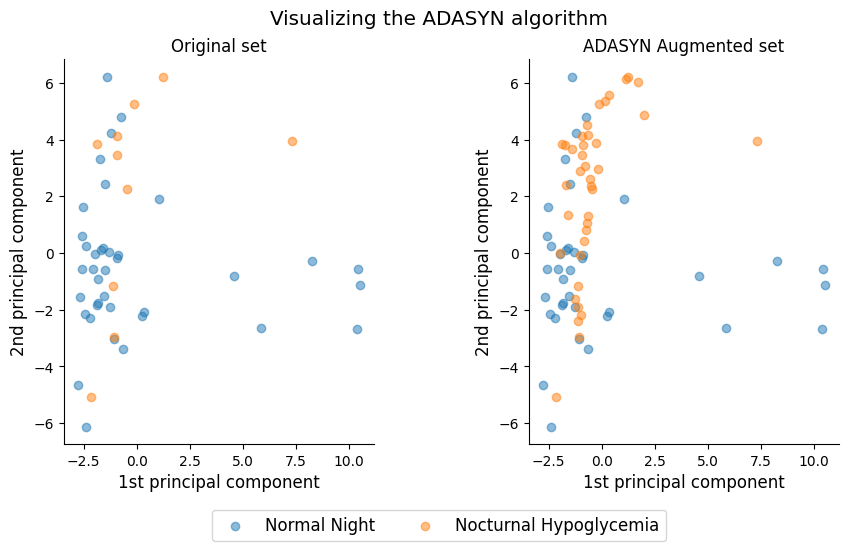}
    \caption{Two component principal component analysis (PCA) representation of the in-house dataset before and after ADASYN \citep{haibohe2008adasyn} augmentation.}
    \label{fig:adasyn}
\end{figure*}

We used the oversampling technique, ADASYN \citep{haibohe2008adasyn} (derived from the SMOTE \citep{Chawla2002SMOTE} algorithm) with a 1-to-1 resampling ratio generating new data points close to the decision boundary. We chose ADASYN to tackle overfitting by generating challenging, hard-to-classify data points. Exposing our models to these tough examples helped them become more robust to outliers and less prone to overfitting. Generating synthetic samples for the minority class balanced our labels and increased our dataset size. In Fig.~\ref{fig:adasyn}, we observe the principal component analysis (PCA) illustration of the in-house dataset between the ADASYN augmented data and the raw data, where the generated samples are close to the decision boundary. This confirms that ADASYN generates hard-to-classify examples, enhancing our models. However, the drawback of oversampling is the introduction of synthetic data points, which can lead to biases in the data. 

\subsection{Feature Engineering}
\label{sec:feat_eng}
Features in the study were categorized into two types: temporal (time-varying) and static (unchanging). We resampled the temporal features to a 15-minute interval. Resampling the data to a 15-minute interval effectively reduces noise while preserving the sensitivity needed to detect critical events in a patient's glucose trajectory, such as cardiovascular variations during and after exertion \citep{battelino2019clinical, barak2010heart}.
The in-house dataset's temporal features consisted of 31 vital signs from the Everion device and 24 static features from logbook data and metadata, totalling 55 raw features for our model. We elaborate on the feature sets in the appendix \ref{apd:feat_set}.


To improve the information extracted from key features like glucose and heart rate, we computed static daily features that capture the daily trends of these variables. This approach is informed by methodologies from previous studies, notably \citet{berikov} and \citet{Sampath2016GlycemicControl}. The description of the functions used to aggregate the readings of the day are listed in Table~\ref{tab:glob_feats}

Data obtained closer to the prediction horizon holds greater significance in our predictive analysis \citep{metcalf2014effects}; hence, on top of full-day trends, we extracted evening trends (7pm to 10pm). Aggregating the temporal dimension and extracting daily features is a technique to remove noise and reduce temporal dimensionality highlighting informative moments of the day.

\begin{table}[h!]
\small
\caption{Calculated daily features and their equations. \(n\) is the number of time steps throughout the day, \(G_i\) represents the glucose level at time step \(i\), \(\bar{G}\) is the mean glucose level, \(\bar{D}\) is the mean of the first differences, \(\sigma_G\) is the standard deviation (SD) of glucose levels, and the "Evening" period refers to a specified time range within the day (7pm - 10pm).}

\begin{center}
\renewcommand*{\arraystretch}{1.5}

\begin{tabular}{cl}
\toprule
\textbf{Function} & \textbf{Equation}
\\ \midrule
Coefficient of Variation & \( \frac{\sigma_G}{\bar{G}} \) \\
Liability Index & \(  \frac{1}{5} \sum_{i=1}^{n-1} (G_{i+1} - G_i)^2 \) \\
SD of the First Differences & \( \sqrt{\frac{\sum_{i=1}^{n-1}(G_{i+1} - G_i - \bar{D})^2}{n-1}} \) \\
Daily Minimum Value & \( \min(G_1, G_2, \ldots, G_n) \) \\
Evening Peak & \( \max(G_{\text{Evening}}) \) \\
Evening Low & \( \min(G_{\text{Evening}}) \) \\
Linear Regression Slope & \( \frac{n \sum_{i=1}^n t_i G_i - \sum_{i=1}^n t_i \sum_{i=1}^n G_i}{n \sum_{i=1}^n t_i^2 - \left( \sum_{i=1}^n t_i \right)^2} \) \\
\bottomrule
\end{tabular}
\label{tab:glob_feats}
\end{center}
\end{table}

We engineered a feature, glucose personalized ($G_{p}$), stemming that individual physiological factors—such as age, height, weight, and BMI—affect the person's glucose metabolism \citep{kashiwagi2023assessment}. To help the model personalise the glucose trends based on a person's physiological profile we designed the following feature, 
\vspace*{-0.1cm}
\begin{equation}
G_{p} = \text{G} \times \left(1 + \left(\text{{a}} + \text{{h}} + \text{{w}} + \text{{b}}\right)\right),
\end{equation}
where $G_{p}$ is the personalised glucose, $G$ is the CGM reading, and a, h, w, b, are the age, height, weight, and BMI of the patient respectively.

Exploring the models' ability to learn on the in-house dataset, we compared performance on multiple feature sets. We analyzed seven feature sets ranging from 50 features (21 temporal, 29 static) to 16 features (6 temporal, 10 static), each chosen to isolate and highlight different aspects of the data:

\begin{description}
    \item[All Features:] Incorporates every available feature to serve as a comprehensive baseline.
    \item[Everion Daily Only:] Focuses on the daily readings from the Everion device to assess the impact of daily aggregated data.
    \item[Glucose Normal:] Uses standard glucose readings, providing a reference for conventional measurements.
    \item[Glucose Personalized:] Adjusts glucose readings based on patient-specific factors to capture personalized insights.
    \item[Non-Aggregated Daily:] Excludes features derived from daily calculations to evaluate the influence of these computed metrics.
    \item[Marx et al. (2023):] Implements the feature set proposed by \citet{alexandre_marx} as a benchmark against established methodologies.
    \item[Reduced Selection:] Employs a refined subset of features chosen based on the feature correlation aimed at optimizing model performance while reducing complexity.
\end{description}

Each feature set was selected with a specific rationale, allowing us to understand the contribution of various data aspects to our predictive performance.

\subsection{Machine Learning Algorithms}
\subsubsection{Base Models}
The Related Work section (Section~\ref{sec:rel_work}) highlighted the robust performance of classical ML models, including the RFC and SVM, over the use of deep neural networks (DNN). These models consistently yield promising results, showcasing their ability to generalize well and mitigate overfitting when data points are limited. 

We use three main models in this work: RFC \citep{rfc_highdim}, Long Short-Term Memory (LSTM) \citep{lstm_d2l} model, and Convolutional Neural Network (CNN) \citep{cnn_def}. 
A RFC is a widely used ML model that aggregates the outputs of multiple decision trees to determine the final predicted class. It is particularly effective in scenarios involving high-dimensional inputs, where its ability to handle complex data makes it a preferred choice \citep{JMLR:v15:delgado14a}.
LSTM and CNN models were also selected as comparative models because they are well-suited to our problem. Both models can handle two-dimensional input, which enables us to separate the time-related data from the other features in the dataset. LSTMs are popular due to their capacity to avoid vanishing gradients \citep{pascanu2012difficulty}, capture uni-directional dependencies over long distances, and generalize effectively to unseen data \citep{greff2016lstm}. A convolutional neural network (CNN) abstracts the problem and excels at local pattern recognition \citep{Alzubaidi2021DeepLearningReview}, making it an ideal exploratory model for assessing the impact of temporal data.

\subsubsection{Daily Variable Models}
The in-house dataset has a mixture of static and temporal features, we maintained this temporal separation by customizing deep learning architectures.  We adjusted the LSTM network and CNN architectures by passing the temporal features through the temporal layers (LSTM and CNN layers) to concatenate their non-temporal vector with our daily variables. For the hidden layers, we pass a Rectified Linear Unit (ReLU) \citep{Agarap2018} as an activation function and for the binary classification a sigmoid \citep{sigmoid_def} activation function. 
Kernel regularization on the dense layers was used to mitigate overfitting. 


\subsubsection{Transfer Learning using OhioT1DM} \label{sec:trans}
Transfer learning involves leveraging knowledge gained from training on a larger, more general dataset (the OhioT1DM dataset) and applying it to a smaller, higher-feature dataset (the in-house dataset). In this process, a model pre-trained on a large dataset applies learned patterns to a new task. Fine-tuning on a smaller dataset allows the model to adapt quickly, needing less data and computational resources than starting from scratch. This approach makes sense for our limited, costly-to-collect dataset, enabling more efficient and effective model training.

The dataset was built from two studies where the participants used different sensor bands. This led to one of the studies not having the temporal physiological readings of the patients. To maximize the data points consistently and still demonstrate the potential of transfer learning, we chose to only extract the glucose recordings of this dataset.

The model used for transfer learning was a neural network consisting of LSTM layers pretrained on the glucose values of the OhioT1DM dataset. We then froze the trained layers and extended the architecture to encompass a set of hand-selected features from the in-house dataset based on the results of the exploratory feature sets. This allows us to leverage the OhioT1DM's cardinality.

The hand-selected features are the following: glucose, hypoglycemic flag, GSR electrode values, activity classification, blood pulse wave, core temperature, heart rate, heart rate variability, motion activity, number of steps, perfusion index, and respiration rate.


\section{Experimental Details}
\paragraph{Implementation}
Our dataset's cardinality remained small throughout the studies; model training was done on CPUs. The libraries used were Numpy \citep{harris2020arraynp}, Pandas \citep{pandas}, Tensorflow \citep{tensorflow2015-whitepaper}, PyTorch \citep{pytorch}, imblearn \citep{imblearn}, and scikit-learn \citep{scikit-learn}. The RFC performed best with 1000 trees and a balancing of class weights. All models ran for 100 epochs. We employed early stopping with a patience of 30 epochs to accommodate the model's highly fluctuating performance during training, ensuring sufficient opportunity for stabilization and convergence. For our models' results, we used Stratified 5-fold Cross Validation. 


\paragraph{Metrics and Evaluation}

Medical datasets often exhibit a significant class label imbalance \citep{imbalance_data_problem, RahmanDavis2013}, which presents a major challenge, especially for conditions like NH. In such cases, the risk of false negatives is particularly critical because failing to predict an NH episode can severely affect patient safety \citep{allen2003nocturnal, danger_hyp}. Standard accuracy measurements fail to adequately address this scenario \citep{imbalanced_data_review}, necessitating careful consideration of metrics.
To mitigate this issue, we utilized the binary cross entropy focal loss, \citep{lin2017focal}, defined as \begin{equation}    
FL(p_t) = -\alpha_t (1 - p_t)^\gamma \log(p_t),
\end{equation} 
with $FL(p_t)$ as the focal loss for a given probability $p_t$, with $p_t$ being the model's estimated probability for the class with the true label $t$, $\alpha_{t}$ as the weight factor for class $t$, helping mitigate class imbalance by assigning more weight to the rare class, $\gamma$ being the focusing parameter that smoothly adjusts the rate at which easy examples are down-weighted, and $\log$ denoting the natural logarithm. We opted for the AUROC score as a critical metric to assess and compare each model's performance.
The AUROC score is a performance measurement for classification problems at various threshold settings, well-defined under imbalanced datasets. It calculates the area under the receiver operating characteristic curve \citep{HajianTilaki2012}. This curve illustrates the performance of a binary classification task along different threshold values. For completeness, we also report the F1-score \begin{equation}
F1 = 2 \times \frac{\text{Precision} \times \text{Recall}}{\text{Precision} + \text{Recall}},\end{equation}
where $\text{Precision} = \frac{TP}{TP + FP}$, with TP as true positives and FP as false positives, and $\text{Recall} = \frac{TP}{TP + FN}$, with FN as false negatives. The F1 score gives a more comparative overview of the models' performances. 

%
\section{Results}
\subsection{In-house: Feature Exploration}
\begin{figure*}[h!]
    \centering
    \includegraphics[width=0.9\textwidth]{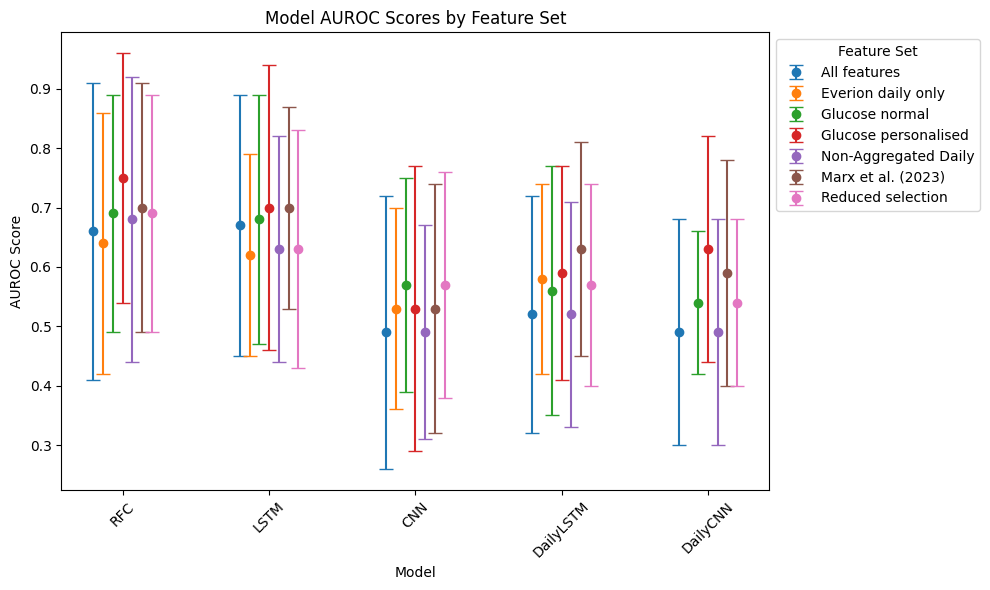}
    \caption{This figure displays the distribution of the mean AUROC cross-validation scores across three different random seeds for the different feature sets across models.}
    \label{fig:resultsPlot}
\end{figure*}

Our initial results, reported in Table \ref{tab:model_results}, show the AUROC performance of the different models over the different feature sets.
Table \ref{tab:f1_diacamp} in Appendix~\ref{apd:third} details the corresponding F1 scores. Figure \ref{fig:resultsPlot} visualizes the distribution of mean AUROC scores across models and feature sets, highlighting performance variations. This provides a more comprehensive understanding of how different feature sets impact predictive performance across models.

\begin{table*}[h!]
\caption{This table compares different feature selection paradigms (Section~\ref{sec:feat_eng}) across different predictive models. Reporting the mean$\pm$standard deviation AUROC scores across three different random seeds, five-fold stratified cross-validation, and ADASYN oversampling on the in-house dataset, for a prediction horizon of 10 pm to 7 am.}
\label{tab:model_results}
\begin{center}
\begin{tabular}{lccccc}
\toprule
\textbf{Feature set} & \textbf{RFC} & \textbf{LSTM} & \textbf{CNN} & \textbf{DailyLSTM} & \textbf{DailyCNN} \\
\midrule 
All features & 0.66 $\pm$ 0.25 & \textbf{0.67 $\pm$ 0.22} & 0.49 $\pm$ 0.23 & 0.52 $\pm$ 0.20 & 0.49 $\pm$ 0.19 \\
Everion daily only & \textbf{0.64 $\pm$ 0.22} & 0.62 $\pm$ 0.17 & 0.53 $\pm$ 0.17 & 0.58 $\pm$ 0.16 & N/A \\
Glucose normal & \textbf{0.69 $\pm$ 0.20 }& 0.68 $\pm$ 0.21 & 0.57 $\pm$ 0.18 & 0.56 $\pm$ 0.21 & 0.54 $\pm$ 0.12 \\
Glucose personalised & \textbf{0.75 $\pm$ 0.21} & 0.70 $\pm$ 0.24 & 0.53 $\pm$ 0.24 & 0.59 $\pm$ 0.18 & 0.63 $\pm$ 0.19 \\
Non-Aggregated Daily & \textbf{0.68 $\pm$ 0.24} & 0.63 $\pm$ 0.19 & 0.49 $\pm$ 0.18 & 0.52 $\pm$ 0.19 & 0.49 $\pm$ 0.19 \\
\citep{alexandre_marx} & 0.70 $\pm$ 0.21 & \textbf{0.70 $\pm$ 0.17} & 0.53 $\pm$ 0.21 & 0.63 $\pm$ 0.18 & 0.59 $\pm$ 0.19 \\
Reduced selection & \textbf{0.69 $\pm$ 0.20} & 0.63 $\pm$ 0.20 & 0.57 $\pm$ 0.19 & 0.57 $\pm$ 0.17 & 0.54 $\pm$ 0.14 \\ 
\bottomrule
\end{tabular}
\end{center}
\footnotesize
Abbreviations: AUROC, Area Under the Receiver Operating Curve; RFC, Random Forest Classifier; LSTM, Long Short Term Memory; CNN, Convolutional Neural Network.
\end{table*}

\vspace*{-0.05cm}

\subsection{OhioT1DM}
\label{sec:in-house_results}
We evaluated our models on the OhioT1DM dataset, reporting the mean AUROC score across three random seeds (see Table \ref{tab:ohiot1dm_model_results}). Because the dataset already had predefined training and test splits, we did not perform stratified cross-validation. To ensure consistency with the in-house data, we selected features that were both reliably recorded and comparable. Specifically, we choose: glucose, hypoglycemia, GSR electrode, basal values, and skin temperature.


\begin{table*}[h!]
\caption{AUROC scores and standard deviation averaged across three different random seeds for the NH prediction on the OhioT1DM dataset with a prediction horizon of 9 hours (10pm - 7am).}
\label{tab:ohiot1dm_model_results}
\begin{center}
    
\begin{tabular}{lccccc}
\toprule
\textbf{Metric} & \textbf{RFC} & \textbf{LSTM} & \textbf{CNN} & \textbf{DailyLSTM} & \textbf{DailyCNN} 
\\ \midrule
AUROC score & 0.60 $\pm$ 0.03 & 0.60 $\pm$ 0.03 & 0.56 $\pm$ 0.03 & 0.47 $\pm$ 0.09 & \textbf{0.67 $\pm$ 0.04} \\ 
\bottomrule
\end{tabular}
\end{center}
\footnotesize
Abbreviations: AUROC, Area Under the Reciever Operating Curve; RFC, Random Forest Classifier; LSTM, Long Short Term Memory; CNN, Convolutional Neural Network.  
\end{table*}

\subsection{Transfer Learning}
After optimally pretraining our model using the OhioT1DM dataset, the cross-validation results on the in-house dataset gave an average AUROC score of \textbf{0.78 $\pm$ 0.05}, using the LSTM described in Section~\ref{sec:trans}. 
\section{Discussion}

\paragraph{Feature Selection and Impact}
Among the evaluated feature sets, the personalized glucose feature set and \citet{alexandre_marx} feature set consistently outperformed the alternative sets. This suggests that a smaller, more informative feature subset enhances model learning. In contrast, using all available features resulted in poorer performance, likely due to noise and redundancy. Furthermore, models restricted to only global features underperformed, highlighting the importance of temporal information. Notably, the consistent drop in AUROC scores when excluding daily computed features further confirms their significance for accurate classification.

\paragraph{Model Performance on the In-House Dataset}
On the in-house dataset, the RFC model using the personalized glucose feature set achieved the highest AUROC score of 0.75 (Table~\ref{tab:model_results}). This indicates that reducing model complexity benefits model performance in such datasets. While both RFC and LSTM models showed competitive AUROC scores across feature sets, more complex models like DailyCNN and DailyLSTM performed worse. Suggesting that simpler architectures can generalize better given the dataset's constraints.


\paragraph{Comparative Dataset Overview}
The in-house and OhioT1DM datasets differ in both scope and study design. The in-house dataset captures children's glucose dynamics in a controlled setting, relying on wearable sensors. In contrast, OhioT1DM includes a broader age range with a greater reliance on self-reported life events. These differences are reflected in model performance, particularly in higher standard deviations for the in-house dataset due to its smaller sample size. Hence, increasing data samples improves convergence and model stability. Additionally, transfer learning with two different datasets still proves valuable in addressing model convergence, despite their dissimilarities in nature.


\paragraph{Cross-Dataset Model Comparison}
Comparing AUROC scores between datasets (Tables~\ref{tab:model_results} and \ref{tab:ohiot1dm_model_results}), key differences emerge. LSTM-based models (LSTM and DailyLSTM) consistently performed better on the in-house dataset. CNN-based models (CNN and DailyCNN) exhibited higher AUROC scores on the OhioT1DM dataset. This discrepancy likely stems from CNN models leveraging larger sample sizes more effectively. Considering how prone the models are to overfitting for this particular dataset and classification task, fine-tuning remains key to achieving the best performance.

\paragraph{Transfer Learning Effectiveness}

Our transfer learning approach yielded the best results, reinforcing its robustness. Despite differences between the datasets, the method achieved an average AUROC of 0.78 (SD: 0.05). This demonstrates low variability and strong predictive performance, indicating that leveraging pre-trained knowledge improves predictive capabilities in varied contexts.

\paragraph{Comparison to Previous Studies}
Despite a smaller sample size (fewer than 314 data points), our findings remain competitive with prior state-of-the-art studies that used significantly larger CGM datasets \citep{lopezCBigDNH, berikov}. While our AUROC scores are slightly lower, our model maintained strong performance using a smaller and more diverse dataset over a broader prediction horizon, which solves a more clinically relevant and challenging problem. Notably, our model outperformed the 3 am - 6 am prediction horizon from \citet{vuExtendedPred2020}, despite having to apply transfer learning from an adult dataset to a children's dataset over an extended 9-hour horizon. This underscores the potential of informative features and transfer learning.

\paragraph{Limitations and Future Works}
The most obvious limitation of this study is the size of the dataset. A small dataset restricts the amount of information available for training and testing the models, leading to overfitting and poor generalizability \citep{Brigato2020ACL}. 
This results in decreased performance when applied to new or unseen data. Despite transfer learning, manually entered logbook entries are inconsistent for both datasets, creating data gaps that need imputation. 
Overall, small sample sizes can introduce uncertainty and inaccuracies, resulting in significantly skewed model results.

For future research, an important focus should be on maintaining the temporal dimensions of our features when oversampling the minority class. One promising approach could be using SMOTE specifically for time-series data \citep{ZhaoTSMOTE2022}, ensuring that the temporal dimension is accounted for during the data augmentation. An important aspect of ML in the medical field is its potential to assist healthcare professionals in making informed decisions. Future research could include confidence intervals in our predictions, which can provide a range of expected values conveying the model's uncertainty.

%
\section{Conclusion}
In conclusion, our work shows successful results in a relevant and critical issue in pediatrics: improving diabetic management for children with T1D, where even small advances can have a major effect. Particularly, we focus on a long prediction horizon, which has a broader impact on the clinics. Using a challenging dataset, due to its size and nature, we experimented with a variety of machine learning techniques, extensive feature engineering, and multiple models to address data complexity. Among these, the best performance on our in-house data was an AUROC score of 0.75 using the personalized glucose feature set and a Random Forest Classifier. Moreover, integrating adult data from a different distribution through transfer learning boosts the average AUROC scores from 0.75 to 0.78, and reduces the standard deviation from 0.21 to 0.05 (a 76\% decrease). This way, we show both positive results in NH prediction and the benefits of incorporating diverse data sources to enhance model robustness and predictive accuracy. These findings pave the way for new research in predictive NH that moves beyond glucose-only methods by incorporating broader physiological data in long prediction horizons. This approach opens the door to more effective and less invasive clinical decision-making tools in pediatrics.

\section*{Acknowledgments}
 SL is supported by the Swiss State
Secretariat for Education, Research, and Innovation (SERI) under contract number MB22.00047.

\newpage

\bibliography{iclr2025_conference}
\bibliographystyle{iclr2025_conference}

\newpage
\appendix
\section{Feature sets}\label{apd:feat_set}

The following subsections contain the features used for the respective feature sets that were compared throughout this study.

\subsection{All Features}
\label{sec:feature_sets}
\textbf{Glucose-related:}\\
Glucose, hypoglycemia event, glucose linear regression slope, glucose evening low and peak, daily glucose minimum, standard deviation of glucose differences, coefficient of glucose variation.

\textbf{Heart-related:}\\
Heart rate, heart rate variability, heart rate evening low and peak, heart rate variability evening low and peak, heart rate variability minimum.

\textbf{Physical Activity:}\\
Motion activity, activity classification, number of steps, perfusion index, respiration rate, energy, activity score, wellness index, evening low and peak.

\textbf{Temperature \& Pressure:}\\
Core temperature, temperature local, temperature object, barometer pressure.

\textbf{Additional Biometrics:}\\
Blood pulse wave, GSR electrode, gender, age, weight, height, BMI, basal percentage, basal total.

\textbf{Insulin \& Diabetes Data:}\\
Glycated hemoglobin (HbA1c) reading, total daily insulin dose (TDD), max daily insulin fast, max daily insulin slow, total daily fast insulin, total daily slow insulin.

\textbf{Others:}\\
Health score, training effect score, richness score.

\subsection{Everion Daily Only}
\textbf{Glucose-related:}\\
Glucose, hypoglycemia flag.

\textbf{Insulin \& Diabetes Data:}\\
Max insulin fast, max insulin slow, total insulin fast, total insulin slow, glycated hemoglobin (HbA1c) reading, total daily insulin dose.

\textbf{Demographics:}\\
Gender, age, weight, height, BMI, basal percentage, basal total.

\textbf{Heart-related:}\\
Heart rate variability evening low and peak, heart rate variability minimum.

\subsection{Glucose Normal}
\textbf{Core Features:}\\
Glucose, hypoglycemia flag, heart rate, heart rate variability, number of steps.

\textbf{Insulin \& Diabetes Data:}\\
Max insulin fast, max insulin slow, total insulin fast, total insulin slow, glycated hemoglobin (HbA1c) reading, total daily insulin dose.

\textbf{Demographics:}\\
Gender, age, weight, height, BMI, basal percentage, basal total.

\textbf{Glucose Metrics:}\\
glucose linear regression slope, glucose evening low, and peak, daily glucose minimum, standard deviation of the glucose differences, coefficient of glucose variation.

\subsection{Personalized Glucose}
\textbf{Glucose Metrics:}\\
Glucose personalised ($G_p$), hypoglycemia events, glucose linear regression slope, glucose evening low and peak, daily glucose minimum, standard deviation of glucose differences, coefficient of glucose variation.

\textbf{Heart-related:}\\
Heart rate, heart rate variability.

\textbf{Insulin-related:}\\
Max insulin fast, max insulin slow, total insulin fast, total insulin slow.

\subsection{Non-Aggregated Daily}
\textbf{Core Features:}\\
Glucose, hypoglycemia flag, heart rate, perfusion index, motion activity, activity classification, heart rate variability, respiration rate, energy, core temperature, temperature local, barometer pressure, GSR electrode, health score, training effect score, activity score, richness score, blood pulse wave, temperature object, temperature barometer.

\textbf{Insulin-related:}\\
Max insulin fast, max insulin slow, total insulin fast, total insulin slow.

\subsection{\citep{alexandre_marx}}
\textbf{Core Features:}\\
Activity classification, blood pulse wave, core temperature, GSR electrode, heart rate, heart rate variability, motion activity, number of steps, perfusion index, respiration rate.

\textbf{Demographics:}\\
Gender, age, weight, height, BMI.

\textbf{Insulin \& Diabetes Data:}\\
Basal percentage, basal total, glycated hemoglobin (HbA1c) reading, total daily insulin dose, max daily insulin fast, max daily insulin slow, total daily fast insulin, total daily slow insulin.

\textbf{Glucose Metrics:}\\
Glucose linear regression slope, glucose evening low and peak, daily glucose minimum, standard deviation of glucose differences, coefficient of glucose variation.

\subsection{Reduced Selection}
\textbf{Core Features:}\\
Glucose, hypoglycemia flag, activity classification, blood pulse wave, core temperature, GSR electrode, heart rate, heart rate variability, motion activity, number of steps, perfusion index, respiration rate.

\textbf{Insulin \& Demographics:}\\
Max daily insulin fast, max daily insulin slow, total daily fast insulin, total daily slow insulin, gender, age, weight, height, BMI, basal percentage, basal total, glycated hemoglobin (HbA1c) reading, total daily insulin dose.

\textbf{Glucose Metrics:}\\
Glucose linear regression slope, glucose evening low and peak, daily glucose minimum, standard deviation of glucose differences, coefficient of glucose variation.

\onecolumn
\section{In-house Recorded Features}\label{apd:feats}

In this appendix section, we list and describe in Tables~\ref{tab:eve_feats} and~\ref{tab:meta_data} all the features recorded from the devices used during the In-house study.

\begin{table}[h!]
    \centering
    
    \caption{List and description of the Everion features, the bold features are the features used in this research.}
\begin{tabular}{ll}
\toprule
\textbf{Everion Feature} & \textbf{Description} \\ 
\midrule
\textbf{Heart rate} & Measures the number of heartbeats per minute.\\ 
Oxygen saturation & Assesses the percentage of oxygen-saturated hemoglobin.\\ 
\textbf{Perfusion index} & Indicates the pulse strength at the sensor site.\\ 
\textbf{Motion activity} & Tracks the movement activity of the wearer.\\ 
\textbf{Activity classification} & Categorizes the type of physical activity performed.\\ 
\textbf{Heart rate variability} & Monitors variations in the time interval between heartbeats.\\ 
\textbf{Respiration rate} & Number of breaths taken.\\ 
\textbf{Energy} & Energy Expenditure.\\ 
\textbf{Core temperature} & Monitors the internal body temperature.\\ 
\textbf{Temperature local} & Temperature at the device’s location on the body.\\ 
\textbf{Barometer pressure} & The atmospheric pressure.\\ 
\textbf{GSR electrode} & Skin’s electrical conductance.\\ 
\textbf{Health score} & Score of the wearer’s health status.\\ 
\textbf{Relax stress intensity score} & Score on the intensity of stress and relaxation levels.\\ 
Sleep quality & index score Score on the quality of sleep. \\ 
\textbf{Training effect score} & Score on the impact of exercise on fitness levels.\\ 
\textbf{Activity score} & Score based on physical activity intensity and duration.\\ 
\textbf{Richness score} & Score based on physical activities undertaken.\\ 
Heart rate quality & Quality of heart rate measurements.\\ 
Oxygen saturation quality & Quality of oxygen saturation measurements.\\ 
\textbf{Blood pulse wave} & Metric of the blood pulse wave.\\ 
\textbf{Number of steps} & Number of steps taken.\\ 
Activity classification quality & Quality of activity classification.\\ 
Energy quality & Quality of energy expenditure estimations.\\ 
Heart rate variability quality & Quality of HRV measurements.\\ 
Respiration rate quality & Quality of respiration rate measurements.\\ 
Core temperature quality & Quality of core temperature measurements.\\ 
\textbf{Temperature object} & Temperature of an object in proximity to the device.\\ 
\textbf{Temperature barometer} & Temperature readings from the device’s built-in barometer. \\ 
Timestamp UTC & Date and time in Coordinated Universal Time.\\ 
Timestamp offset & Local time offset from UTC at the time of measurement.\\ 
\bottomrule
\end{tabular}
    
\label{tab:eve_feats}
\end{table}

\begin{table*}[h!]
    \centering
    \caption{List and description of metadata features.}
    \begin{tabular}{p{5cm}p{8cm}}
    \toprule
    \textbf{Metadata Feature} & \textbf{Description} \\
    \midrule
Gender & The biological sex of the individual (male or female).
\\
Age & The age of the individual is typically measured in years.
\\
Weight & The body weight of the individual, usually measured in kilograms (kg) or pounds (lbs).
\\
Height & The stature of the individual, typically measured in centimetres (cm) or inches (in).
\\
BMI (Body Mass Index) & A measure of body fat based on height and weight. It is calculated as weight in kilograms divided by the square of height in meters (kg/m²).
\\
Basal Percentage & The percentage of total daily insulin that is basal (background insulin) to manage glucose levels over time.
\\
Basal Total & The total daily amount of basal (long-acting) insulin, typically measured in units.
\\
HbA1c (Glycated Hemoglobin) & A measure of average blood glucose levels over the past 2-3 months. It is expressed as a percentage and used to monitor long-term glucose control in people with diabetes.
\\
TDD (Total Daily Insulin Dose) & The total amount of insulin taken in a day, including both basal (long-acting) and bolus (fast-acting) insulin, typically measured in units.
       \\  
       \bottomrule
    \end{tabular}
    \label{tab:meta_data}
\end{table*}

\vspace*{5cm}
\section{Additional Results}\label{apd:third}
We show in Table~\ref{tab:f1_diacamp} the spread of the F1 scores for the different proposed feature sets and models applied to the In-house dataset. This table is used as an extension of Section~\ref{sec:in-house_results} that contains the corresponding
AUROC mean averages.

\begin{table*}[h!]
\centering
\caption{
Mean$\pm$standard deviation F1 scores across three different random seeds, five-fold
stratified cross-validation, and ADASYN oversampling on the in-house dataset, for a prediction
horizon of 10 pm to 7 am. This table compares different feature selection paradigms (Section~\ref{sec:feat_eng}) across different predictive models.}

\begin{tabular}{lccccc}
\toprule
\textbf{Feature set} & \textbf{RFC} & \textbf{LSTM} & \textbf{CNN} & \textbf{DailyLSTM} & \textbf{DailyCNN} \\
\midrule
All features &\textbf{ 0.43 $\pm$ 0.30} & 0.43 $\pm$ 0.18 & 0.34 $\pm$ 0.18 & 0.25 $\pm$ 0.22 & 0.26 $\pm$ 0.14 \\
Everion daily only & \textbf{0.47 $\pm$ 0.28} & 0.45 $\pm$ 0.18 & 0.20 $\pm$ 0.21 & 0.30 $\pm$ 0.22 & N/A \\
Glucose normal & \textbf{0.49 $\pm$ 0.28} & 0.41 $\pm$ 0.21 & 0.32 $\pm$ 0.18 & 0.20 $\pm$ 0.18 & 0.29 $\pm$ 0.16 \\
Glucose personalise & \textbf{0.54 $\pm$ 0.26} & 0.45 $\pm$ 0.22 & 0.25 $\pm$ 0.25 & 0.16 $\pm$ 0.18 & 0.28 $\pm$ 0.24 \\
Non-Aggregated Daily & \textbf{0.48 $\pm$ 0.32} & 0.31 $\pm$ 0.18 & 0.29 $\pm$ 0.18 & 0.24 $\pm$ 0.18 & 0.08 $\pm$ 0.18 \\
\citep{alexandre_marx} & \textbf{0.52 $\pm$ 0.32} & 0.41 $\pm$ 0.21 & 0.28 $\pm$ 0.19 & 0.33 $\pm$ 0.17 & 0.28 $\pm$ 0.25 \\
Reduced selection & \textbf{0.54 $\pm$ 0.31} & 0.42 $\pm$ 0.17 & 0.28 $\pm$ 0.18 & 0.26 $\pm$ 0.24 & 0.18 $\pm$ 0.18 \\
\bottomrule
\end{tabular}
\footnotesize
\newline
\newline
Abbreviations: RFC, Random Forest Classifier; LSTM, Long Short Term Memory; GV, Daily Variable; REG, Regularised; CNN, Convolutional Neural Network.  
\label{tab:f1_diacamp}
\end{table*}

\end{document}